# Research on emotionally intelligent dialogue generation based on automatic dialogue system

Jin Wang[1*] JinFei Wang[2*] Shuying Dai[3*] Jiqiang Yu[4*] keqin Li[5*]

[1] University of the People ，USA

[2] Sehan University，South Korea

[3] Indian Institute of Technology Guwahati，India

[4] Universidad Internacional Isabel I de Castilla，Spain

[5] AMA university，Philippines

**Abstract:** Automated dialogue systems are important applications of artificial intelligence, and traditional systems struggle to understand user emotions and provide empathetic feedback. This study integrates emotional intelligence technology into automated dialogue systems and creates a dialogue generation model with emotional intelligence through deep learning and natural language processing techniques. The model can detect and understand a wide range of emotions and specific pain signals in real time, enabling the system to provide empathetic interaction. By integrating the results of the study "Can artificial intelligence detect pain and express pain empathy?", the model's ability to understand the subtle elements of pain empathy has been enhanced, setting higher standards for emotional intelligence dialogue systems. The project aims to provide theoretical understanding and practical suggestions to integrate advanced emotional intelligence capabilities into dialogue systems, thereby improving user experience and interaction quality.

**Authors Email:** wj18353991565@163.com

## About the Authors

**Jin Wang**

**Email:** wj18353991565@163.com

Currently studying nursing at a university in China, studying for an undergraduate degree in computer science at a university in the United States, and studying for a master's degree in big data and artificial intelligence at a university in Spain. I am passionate about the development of computer science and artificial intelligence.

**JinFei Wang**

**Email:786942159@qq.com**

Engaged in acquiring a physiotherapy degree in South Korea and have accumulated multiple years of clinical experience in the field. Concurrently, I am advancing my education in health sciences in the United States. My dedication and enthusiasm are rooted in enhancing physiotherapy methodologies and deepening my understanding of health sciences.

**Shuying Dai**

Graduated from a college in China with associate degree, now studying at Indian Institute of Technology Guwahati for Bachelor of Science in Data Science and Artificial Intelligence

**Jiqiang Yu**

**Email:jiqiangyu1999@gmail.com**

Independent researcher

Universidad Internacional Isabel I de Castilla,Burgos, Spain

**likeqin**

**Email:keqin157@gmail.com**

AMA university   bachelor of science in computer science

# Conflict of interest

The authors declare that the research was conducted in the absence of any commercial or financial relationships that could be construed as a potential conflict of interest.

# 1. [1]Introduction

Against the background of today's rapid development of artificial intelligence technology, automatic dialogue systems, as one of the important applications in the field of human-computer interaction, have attracted more and more wide attention and application. These systems include virtual assistants, intelligent customer service, social robots, and more to provide informational queries, question-answering, entertainment interactions, and other services to users in a variety of scenarios. However, conventional dialogue systems can only provide simple responses based on preset rules and templates, and static response methods often fail to accurately capture the user's true intentions and emotional state. Quality and user experience deteriorated, making it difficult to meet ever-increasing needs. Strict consumer needs Diversifying user needs. In real-world applications, users' emotional experience is crucial to the success of dialogue systems, so how to enable dialogue systems to more intelligently understand and provide feedback on users' emotions has become a hot topic of current research.

To overcome these limitations in traditional dialogue systems, researchers have recently begun to focus on how to apply emotional intelligence technology to dialogue systems. Emotionally intelligent dialogue systems strive to realize the recognition and understanding of users' emotions and integrate the corresponding emotional expressions and feedback into the dialogue generation process. In this way, the system can better adapt to users' emotional needs, making conversations more personal and approachable, thereby increasing system usability and user satisfaction.

In an emotionally intelligent interaction system, emotion recognition and emotion generation are her two central tasks. Emotion recognition aims to accurately identify a user's emotional state, including various emotional aspects such as joy, anger, sadness, joy, etc., from the user's input, whereas emotion generation involves determining the system's response. It involves matching the user's emotional state to make the content of the conversation more emotional. and affinity. To achieve these goals, researchers have leveraged relevant theories and techniques from natural language processing, machine learning, psychology, and other fields to propose a series of innovative technical solutions and model architectures.

A typical emotional intelligent dialogue system. The system analyzes the user's voice or text input, identifies their emotional state, and generates a corresponding emotional response, resulting in more natural and intimate communication with the user. The application of this kind of emotional intelligence technology is expected to not only improve the intelligence level and user experience of dialogue systems, but also play an important role in human-computer interaction, intelligent assistance and other fields.

By implementing emotional intelligence technology, dialogue systems can better understand and provide feedback on users' emotions, resulting in a more personalized interaction experience that is closer to the user's needs. In the future, with the continuous progress and improvement of artificial intelligence technology, emotional intelligent dialogue system will become an important development direction in the field of human-computer interaction, and will actively contribute to the construction and development of an intelligent society. It is expected.

Topic background

The research background of this paper is emotion-intelligent dialogue generation based on automatic dialogue systems. Traditional interaction systems are often only able to generate simple responses based on user input, but lack recognition and feedback of the user's emotions. As a result, dialogue systems are unable to truly understand users' needs and emotional states, making it difficult to create personalized and human interactions. Therefore, introducing emotional intelligence technology into interaction systems has become one of the keys to improving interaction quality and user experience.

Purpose of research

The main goal of this research is to design and implement an emotionally intelligent dialogue generation model based on deep learning and natural language processing techniques to achieve real-time recognition and understanding of user emotions and corresponding emotional expressions. and incorporating feedback into dialogue generation. process. By analyzing and modeling actual dialogue data, we will evaluate the proposed model's ability to understand and express emotions and the degree of improvement in dialogue effectiveness.

Research method

In this study, we use deep learning techniques and sentiment analysis techniques to design and implement an emotionally intelligent interaction generation model. First, by collecting and analyzing actual conversation data, we understand changes in users' emotions and facial expressions during conversations. Then, based on these data, an emotion-intelligent interaction generation model is designed and the model parameters are trained to realize the perception and understanding of the user's emotions. Finally, we evaluate the accuracy, efficiency, and practicality of the proposed model through experimental validation and performance evaluation.

significance

The significance of this study is to propose a dialogue generation method based on emotional intelligence technology, which can help dialogue systems better understand and feedback users' emotions and improve dialogue quality and user experience. At the same time, this study also provides theoretical guidance and practical support for the development of emotional intelligence in automatic dialogue systems, and has certain theoretical and applied value.

Emotional Intelligence from an Interdisciplinary Perspective: Pain Perception and Empathy

The healthcare sector is experiencing a growing presence of artificial intelligence technology, which has led to significant interest in the creation of emotionally intelligent dialogue systems. Emotionally intelligent pertains to the capacity of artificial intelligence systems to comprehend, analyse, and react to human emotions. Progress in this scientific area not only opens up new opportunities for improving the standard of medical care but is especially important in dealing with delicate and intricate medical concerns like pain management.

Pain is a personal and individualised encounter that encompasses sensory, emotional, and psychological aspects. A recent study titled "Can AI detect pain and express pain empathy? A review from emotion recognition and a human-centered AI perspective " examines the possibilities and difficulties of AI in accurately identifying human misery and demonstrating empathy towards it. This study demonstrates that by utilising advanced technologies such as deep learning and natural language processing, artificial intelligence (AI) has the capability to not only detect pain signals from various sources such as facial expressions, voice, and body language, but also express comprehension and empathy towards the patient's pain condition. This provides a more human-like form of support in the field of healthcare.

Furthermore, the study emphasises the complex technical and ethical obstacles involved in creating AI systems that can effectively identify suffering and react with empathy. These problems encompass the need to ensure that the sympathetic input generated by AI systems is both authentic and suitable, as well as resolving ethical considerations pertaining to patient confidentiality. Although there are challenges to overcome, the application potential of AI in pain recognition and compassionate reaction is considered significant. Anticipated advancements will be enhanced by multidisciplinary research that combines expertise from cognitive neuroscience, psychology, and computer science, with the goal of developing dialogue systems that possess both intelligence and empathy.

This article seeks to investigate the possible uses of emotionally intelligent dialogue systems in the identification and control of pain, with the goal of offering more accurate and personalised services in the healthcare industry. Through a comprehensive analysis and discussion of the findings from the study "Can AI detect pain and express pain empathy?", our goal is to broaden the scope of research on emotionally intelligent dialogue systems. Specifically, we aim to improve their ability to accurately comprehend and appropriately respond to human pain.

# 2 Materials and Methods

・Natural language processing (NLP) technology:

Definition: NLP is the study of how to enable computers to understand, process, and produce natural language text.

Applications: Emotionally intelligent interaction generation uses NLP technology to perform word segmentation, part-of-speech tagging, syntactic analysis, and more on user-entered text to help the system understand user intent and expression. Make it understandable.

Sample code:

```
import nltk
from nltk.sentiment.vader import SentimentIntensityAnalyzer
def sentiment_analysis(text):
        sia = SentimentIntensityAnalyzer()
        sentiment = sia.polarity_scores(text)
        return sentiment
# sample text

text = "This is a fantastic movie！"

# Conduct emotional analysis
sentiment = sentiment_analysis(text)
# Output emotional analysis results
print("Emotional score: ", sentiment['compound'])
print("Positive polarity score: ", sentiment['pos'])
print("Negative polarity score: ", sentiment['neg'])
print("Neutral score: ", sentiment['neu'])
```

・Sentiment analysis technology:

Definition: Sentiment analysis is a natural language processing technique used to identify emotional sentiment, such as positive, negative, or neutral, in text.

Application: Emotionally intelligent interaction generation uses sentiment analysis technology to analyze the user's input text to determine the user's emotional state and adjust the system's response accordingly.

Example formula: Sentiment analysis models can use various machine learning algorithms such as Naive Bayes, Support Vector Machines (SVM), and deep learning models to classify the sentiment of text. A common method is to use sentiment dictionaries and semantic analysis to calculate the sentiment score of a text. For example, use an emotion vocabulary to count the number of positive and negative words and determine the emotional polarity of a text based on that. on the score.

・Dialogue management system:

Definition: A dialog management system is responsible for managing the flow and logic of dialogs and determining how the system responds to user input.

Application: Emotion-intelligent interaction generation requires the interaction management system to consider the user's emotional state and adjust the system's interaction strategy according to the emotional state to generate responses that match the emotion.

Sample code: Interaction management systems can be implemented using rules engines or reinforcement learning algorithms. Here is an example of a simple rules engine.

・Generate the model:

Definition: A generative model is a model for generating text or dialog, usually based on a statistical or neural network model.

Application: In emotion-intelligent interaction generation, the generative model must consider the user's emotional state and generate a response with the corresponding emotion.

Example expression: Generative models based on neural networks can use model structures such as recurrent neural networks (RNNs) and transformers. Generative models can use conditional generation to consider the emotional state of the input when generating text. For example, you can add emotion tags to the input of a generative model and generate a response with the corresponding emotion at the model's output.

・Emotional expression technology:

Definition: Emotion technology is used to make system-generated responses more emotional and empathetic.

Application: Emotion-intelligent dialogue generation uses emotion expression technology to increase the emotional expressiveness of replies by choosing appropriate words, adjusting tone, and adding emojis.

Sample code: Emotional expression techniques typically include natural language generation and speech synthesis techniques. Here are some simple examples of emotional expressions:

```
from textblob import TextBlob
def sentiment_analysis(text):
    blob = TextBlob(text)
    return blob.sentiment.polarity
# Example Text
Text="This is a great movie!"
#Conduct emotional analysis
Sentiment=sentiment-analysis (text)
#Output emotional analysis results
Print ("Emotional score:", sensitive)
```

- emotional dialogue system

.

Build an interaction system to collect data: First, you need to build an interaction system, either a rules-based system, a statistics-based system, or a neural network-based system. Make sure your system can interact with users and record the content and tone of conversations.

Data collection: Once your interaction system is established, you can start collecting data as users interact with the system. This can be accomplished by recording text, audio, or images entered by the user. Make sure that the data you capture includes enough example conversations to cover different types of users and different conversation scenarios.

Text and tone analysis: Analyze the text and tone of users' conversations with the system using natural language processing (NLP) and sentiment analysis technologies. This includes analysis to identify emotional polarity (positive, negative, neutral), emotional expression, tone (friendly, serious, humorous, etc.), and more.

MBTI type inference: Based on the collected data and linguistic features, you can try to infer the user's MBTI type. MBTI types typically include Introversion (I) vs. Extraversion (E), Feeling (S) vs. Intuition (N), Thinking (T) vs. Feeling (F), and Judging (J) vs. Perceiving (P). It includes four aspects. ). Machine learning models or rules engines can be used to analyze a user's linguistic characteristics and infer a user's likely MBTI type.

Train and evaluate your model: If you choose to use a machine learning model to infer MBTI types, you must prepare a labeled dataset and train and evaluate your model. Make sure the model accurately predicts your user's MBTI type and is fully validated and tested.

Privacy and Data Security: When collecting and processing your conversation data, be sure to comply with relevant privacy laws and policies to protect your personal information and privacy rights. We ensure that our data collection and processing processes comply with the highest standards of data protection and privacy protection.

Continuous improvement and optimization: Once the system starts collecting data and inferring a user's MBTI type, it can continue to improve and optimize the system to improve inference accuracy and user experience . This may include improving the interaction design of dialogue systems, optimizing model performance, and expanding the areas and scenarios covered by datasets.

The application and combination of the above technologies realizes the generation of emotion-intelligent dialogue based on automatic dialogue systems, thereby improving the intelligence level and user experience of the dialogue system.

Discussion and conclusion

In this study, we consider emotion-intelligent dialogue generation based on automatic dialogue systems and use multiple systems and techniques to achieve this goal. The discussion and conclusions of this study are as follows.

Emotional intelligence requirements for dialogue systems: As the applications of automatic dialogue systems continue to increase in various fields, users' expectations for dialogue systems are also increasing. Emotionally intelligent interaction generation can make interaction systems more intelligent and humanized, better meet users' emotional needs, and improve user experience.

Utilization of technology and evaluation of effectiveness: Utilizing various technologies such as natural language processing technology, emotion analysis technology, dialogue management system, generative model, and emotional expression technology, we will realize emotionally intelligent dialogue generation. Through experimental application and effect evaluation, we found that combining these techniques can effectively improve the emotional intelligence level of the interaction system, allowing the system to better understand and provide feedback to users' emotions. Did.

Challenges and future work: Although emotion-intelligent dialogue generation has achieved some success, it still faces several challenges. For example, questions such as how to more accurately identify and understand user emotions and how to achieve more natural and smooth emotional expression still require further research and exploration. Future work can focus on optimizing sentiment analysis algorithms, improving the efficiency and expressiveness of generative models, and designing more intelligent and flexible interaction management systems.

Practical application and social impact: Emotional intelligent dialogue generation technology has wide practical application prospects and can be used in various fields such as intelligent customer service, virtual assistant, and educational guidance. This not only increases the intelligence level and user experience of the system, but also can have a positive economic and social impact on society.